# Artificial Immune Tissue using Self-Organizing Networks

Jan Feyereisl⋆ and Uwe Aickelin⋆

⋆School of Computer Science
University Of Nottingham, Nottingham,
NG8 1BB, UK
`jqf,uxa@cs.nott.ac.uk`

**Abstract**

As introduced by Bentley et al. (2005), artificial immune systems (AIS) are lacking tissue, which is present in one form or another in all living multi-cellular organisms. Some have argued that this concept in the context of AIS brings little novelty to the already saturated field of the immune inspired computational research. This article aims to show that such a component of an AIS has the potential to bring an advantage to a data processing algorithm in terms of data pre-processing, clustering and extraction of features desired by the immune inspired system. The proposed tissue algorithm is based on self-organizing networks, such as self-organizing maps (SOM) developed by Kohonen (1996) and an analogy of the so called Toll-Like Receptors (TLR) affecting the activation function of the clusters developed by the SOM.

## 1 Introduction

A number of immune inspired systems have been developed over the years. From negative selection based algorithms to the self vs. non-self (Forrest et al., 1996) and the danger model (Aickelin et al., 2003). Bentley et al. (2005) argue that tissue is one missing component of AIS, as it is the first line of defence against viruses and bacteria, which possibly initiates the activity of the whole immune system.

### 1.1 Tissue

Tissue is any part of a multi-cellular organism, which provides an environment, that can be affected by viruses and bacteria and thus initiate an immune response. It is an intermediate layer between a problem and the actual immune system, which provides a certain interpretation of the occurring problem to the AIS in order to better protect itself.

### 1.2 TLRs

TLRs are a set of receptors on the surface of immune cells, such as dendritic cells, which act as sensors to foreign microbial products essential to their existence. When encountering one or more of such products, they trigger a cascade of events potentially resulting in an immune response. Different combinations of activated TLRs perform different actions.

## 2 SOM and Intrusion Detection

SOMs have been used as part of an IDS on a number of occasions, nevertheless their main application so far has been in the area of network packet analysis. Our proposed method looks at the use of the SOM algorithm in a number of distinctly different ways. Firstly, the SOM algorithm is only a part of an overall tissue algorithm comprising of a set of functions analogous to biologically real tissue, e.g. the notion of inflammation, TLRs, antigens, etc... Secondly, the aim of an artificial tissue is not to act as an IDS on its own, but rather as an initial pre-processing of system data. Thus it supplies the AIS with 'interesting data', making it easier, quicker and more reliable for the AIS to make a decision about a potential threat to the system. As in the human body, the artificial tissue is an environment in which the initial interactions and alarms are raised when 'something' is happening.

## 3 The Link

There are four main areas of the biological analogy; Tissue, cells, TLRs and inflammation. A general overview of the proposed algorithm design can be seen in Figure 1.

### 3.1 Artificial Tissue

Tissue is a layer between the problem and the AIS, represented in terms of a pre-processing algorithm. It is an environment, within which malignant organisms (i.e. malicious code) invade cells in order to survive and eventually cause damage. In this way, tissue acts as an encoding and reduction layer for the incoming data into the AIS. It analyses the data based on an immunological concept and only passes the 'interesting' data to the AIS. By 'interesting', we mean data which is of potentially unknown nature to the tissue environment. Tissue can be seen as a grid of neurons within a SOM.

### 3.2 Artificial Cells

Tissue comprises of cells, each of which might have slightly different functionality. We can imagine an artificial cell in terms of a neuron within a self organizing map. This means, that a cell has a number of inputs, which are used to compare the incoming data to the tissue to the data that the cell holds, in order to find the most suitable cell to which to relate. Such a cell comes in contact with data that is similar to the cells' content and is eventually adjusted, as well as its neighbouring cells, according to the incoming data, based on the SOM algorithm. The outcome of this automatic cell 'growth' results in the tissue being compartmentalized according to similar types of cells, based on the correlation of the multidimensional input features incoming into the tissue. In other words, similar system behaviour is grouped together within the tissue. This results in a constantly updated map, which holds information about the normal behaviour of a system as a whole. Once an unusual action occurs, this should affect cells within the tissue, that have not been affected before or that have not been affected in such a dramatic way.

### 3.3 Artificial TLRs

The analogy of TLRs is based on the enhancement of the functionality of the tissue cells described above. In the immune system, TLRs sense specific predefined chemicals, which are released by malignant organisms. In a similar way, we can specify a set of potentially hazardous system features, each of which can be represented as a receptor. The TLRs will be associated with the cells within the tissue, as in real life, and based on their activation, they will affect the 'growth' of the cell in a more dramatic way. Similarly to the natural functionality of TLRs, the artificial receptors will have a different impact on the underlying cell if a combination of them are activated at the same time.

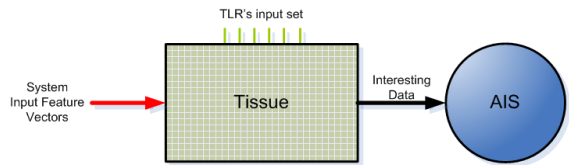

Figure 1: SOM Tissue Design, Cells represented as intersections of white lines, Inflammation as the bandwidth of the tissue I/O streams

### 3.4 Artificial Inflammation

Inflammation proposes the possibility of signalling where the AIS should possibly focus its attention on, or where priority is to be set, thus possibly enabling the notion of problem locality. For example as a result of a rapid cell 'growth', the system can increase or decrease the priority of an associated process. Similarly a technique described by Somayaji and Forrest (2000) can be used in order to give the AIS a better chance at making a correct decision.

## Acknowledgments

This research is partially funded by the ARTIST network (EPSRC GR/S56621/01).